\documentclass[lettersize,journal]{IEEEtran}
\usepackage{amsmath,amsfonts}
\usepackage{algorithm}
\usepackage{algpseudocode}
\usepackage{array}
\usepackage{bm}
\usepackage{textcomp}
\usepackage{booktabs}
\usepackage{stfloats}
\usepackage{multirow}
\usepackage{verbatim}
\usepackage{subcaption}
\usepackage{graphicx}
\usepackage{multirow}
\usepackage{booktabs}
\usepackage{adjustbox}
\usepackage{comment}
\usepackage{hyperref}
\usepackage{amssymb}
\usepackage{breakurl}  
\usepackage{color} 
\usepackage{xcolor}
\usepackage{todonotes}

\usepackage[font=small]{caption}
\usepackage[ defaultcolor=black,commandnameprefix=ifneeded]{changes}
\usepackage{pifont} 
\usepackage{xspace}
\usepackage{cite}
\newcommand{\M}{{\selectfont \textbf{SADER}}\xspace}
\newcommand{\dd}{\mathrm{d}}

\begin{document}
\title{\M: \textbf{S}tructure-\textbf{A}ware Diffusion Framework with \textbf{DE}terministic \textbf{R}esampling for Multi-Temporal Remote Sensing Cloud Removal}
\author{Yifan Zhang$^*$,  Qian~Chen$^*$, Yi Liu$^*$, Wengen Li$^\#$,~\IEEEmembership{Member,~IEEE}, Jihong Guan
\thanks{This work has been submitted to the IEEE for possible publication. Copyright may be transferred without notice, after which this version may no longer be accessible.}
\thanks{Yifan Zhang is with the College of Literature, Science, and the Arts, University of Michigan, Ann Arbor, Michigan, 48105, USA (e-mail: \href{mailto:yifanzhg@umich.edu}{yifanzhg@umich.edu}.}
\thanks{Qian~Chen, Yi~Liu, Wengen Li and Jihong Guan are with the School of Computer Science and Technology, Tongji
University, Shanghai 200092, China (e-mail: \href{mailto:2250951@tongji.edu.cn}{2250951@tongji.edu.cn}; \href{mailto:liuyi61@tongji.edu.cn}{liuyi61@tongji.edu.cn};
\href{mailto:lwengen@tongji.edu.cn}{lwengen@tongji.edu.cn}; \href{mailto:jhguan@tongji.edu.cn}{jhguan@tongji.edu.cn}).}
\thanks{$^*$ Equal contribution.}%
\thanks{$^\#$ Corresponding author.}%
}

\markboth{IEEE Transactions on Knowledge and Data Engineering,~Vol.~xx, No.~x, xxx~2026}%
{Shell \MakeLowercase{\textit{et al.}}: Bare Demo of IEEEtran.cls for Journals}

\maketitle

\begin{abstract}
Cloud contamination severely degrades the usability of remote sensing imagery and poses a fundamental challenge for downstream Earth observation tasks. Recently, diffusion-based models have emerged as a dominant paradigm for remote sensing cloud removal due to their strong generative capability and stable optimization. However, existing diffusion-based approaches often suffer from limited sampling efficiency and insufficient exploitation of structural and temporal priors in multi-temporal remote sensing scenarios. In this work, we propose \M, a structure-aware diffusion framework for multi-temporal remote sensing cloud removal. \M~ first develops a scalable Multi-Temporal Conditional Diffusion Network (MTCDN) to fully capture multi-temporal and multimodal correlations via temporal fusion and hybrid attention. Then, a cloud-aware attention loss is introduced to emphasize cloud-dominated regions by accounting for cloud thickness and brightness discrepancies. In addition, a deterministic resampling strategy is designed for continuous diffusion models to iteratively refine samples under fixed sampling steps by replacing outliers through guided correction. Extensive experiments on multiple multi-temporal datasets demonstrate that \M~ consistently outperforms state-of-the-art cloud removal methods across all evaluation metrics. The code of \M is publicly available at \url{https://github.com/zyfzs0/SADER}.

\end{abstract}


\IEEEpeerreviewmaketitle

\section{Introduction}

\begin{figure}[t]
    \centering
    \includegraphics[width=\linewidth]{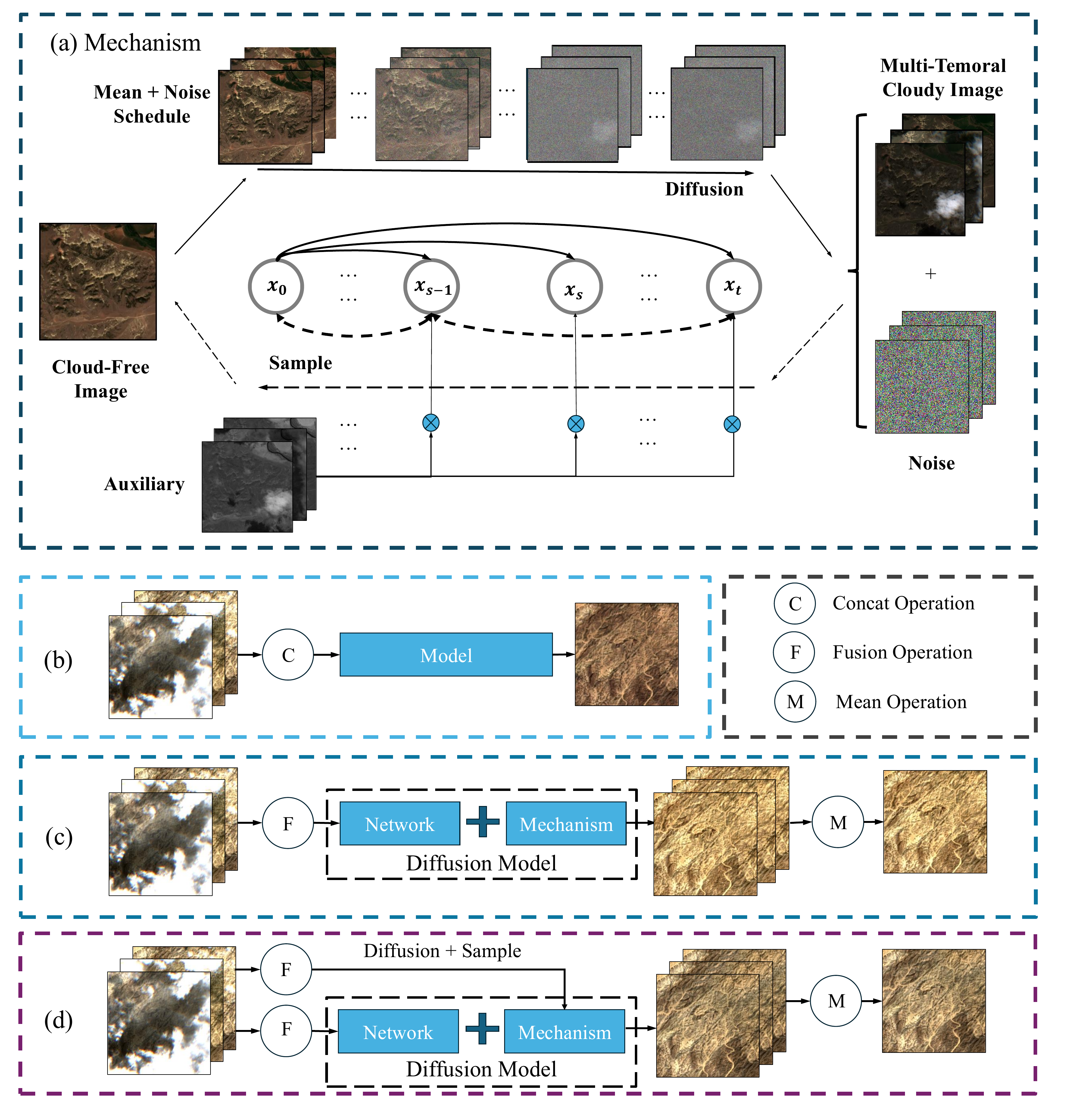}
    \caption{Multi-temporal conditional diffusion for remote sensing cloud removal.
    (a) The diffusion mechanism, illustrating the iterative noise injection and sampling process driven by a mean--noise schedule.
    (b) Prior methods that incorporate multi-temporal information via simple input concatenation before a single predictive model.
    (c) Latest diffusion method (EMRDM) that integrates multi-temporal cues into the network, while the diffusion mechanism remains fixed.
    (d) The proposed \M framework, which incorporates multi-temporal information into both the network and the diffusion mechanism.
    Here, the diffusion model couples a network for step-wise prediction with a diffusion mechanism that iteratively refines the reconstruction.}
    \label{fig:stria_framework}
    \vspace{-2ex}
\end{figure}

\IEEEPARstart{R}{emote} sensing imagery is a vital means of acquiring surface information and has been widely applied in environmental monitoring~\cite{faisal2012remote}, agricultural management~\cite{huang2018agricultural}, economic assessment~\cite{li2017economic}, urban expansion, etc.
However, cloud coverage remains a major obstacle to the acquisition and effective utilization of remote sensing data.
According to the International Satellite Cloud Climatology Project (ISCCP)~\cite{schiffer1983international}, approximately 67\% of the Earth’s surface is affected by clouds at any given time.
Cloud contamination obscures or distorts surface observations, significantly reducing image clarity and reliability~\cite{wang2024deep}, and consequently degrades the performance of downstream tasks such as land cover classification, object detection, and change monitoring~\cite{jing2023denoising}.
Therefore, developing effective cloud removal methods to reconstruct cloud-free remote sensing imagery is of great importance for improving data usability and supporting reliable geospatial analysis.

From a problem formulation perspective, remote sensing cloud removal can generally be categorized into two paradigms: mono-temporal and multi-temporal cloud removal~\cite{zhu2017deep}.
Mono-temporal cloud removal relies solely on a single cloud-contaminated observation to reconstruct the underlying surface, where the available information is inherently limited and strong priors or generative assumptions are often required.
In contrast, multi-temporal cloud removal incorporates multiple observations of the same region acquired at different times as conditional information, exploiting cross-temporal structural consistency and information redundancy to provide richer constraints for reconstruction~\cite{sarukkai2020cloud}.
However, multi-temporal observations also introduce temporal variations and observation inconsistency, making it challenging to effectively model and fully exploit multi-temporal conditional information for robust cloud removal~\cite{jing2023denoising}.

Existing cloud removal methods for remote sensing imagery have been largely driven by advances in computer vision and signal processing~\cite{zhu2017deep, li2025diffusion}.
Early approaches rely on physical modeling and handcrafted priors to exploit spectral and temporal correlations, but generally yield limited reconstruction fidelity~\cite{huang2022ctgan}.
With the development of deep learning, data-driven methods such as autoencoders and GANs have been widely adopted, learning direct mappings from cloud-contaminated inputs to cloud-free outputs~\cite{sintarasirikulchai2018multi, sarukkai2020cloud}.
While these approaches improve visual quality to some extent, they largely follow a point-estimation reconstruction paradigm and lack explicit mechanisms to model spatial-temporal uncertainty, which limits their ability to fully exploit reliability differences and structural redundancy in multi-temporal observations.

More recently, diffusion models have been introduced into remote sensing cloud removal and have demonstrated promising performance~\cite{zhao2023cloud, zou2024diffcr}.
By formulating cloud removal as an iterative denoising process, diffusion-based approaches provide a principled generative framework with stable optimization behavior and strong representation capacity.
However, when directly applied to high-resolution remote sensing imagery, conventional diffusion models often require a large number of denoising steps and may exhibit limitations in preserving fine-grained structures~\cite{liu2025effective}.

Beyond these general limitations, multi-temporal remote sensing cloud removal presents challenges that remain insufficiently addressed by existing diffusion-based methods.
Although multi-temporal observations provide valuable temporal redundancy and structural cues, most approaches incorporate them only as static conditional inputs, without fully and coherently integrating such information into the diffusion modeling and sampling process.
Consequently, temporal priors and structural consistency across multiple acquisitions are not effectively reflected in the diffusion dynamics, limiting the exploitation of complementary information across time.

To address these challenges, we propose \M, a \textbf{S}tructure-\textbf{A}ware diffusion framework with \textbf{DE}terministic \textbf{R}esampling for multi-temporal remote sensing cloud removal.
Inspired by recent mean-reverting diffusion models~\cite{liu2025effective}, \M integrates structure-aware design into both the diffusion modeling and sampling process, enabling more effective exploitation of structural consistency. Specifically, structure-aware information is integrated at the network, loss, and sampling levels to guide the denoising dynamics toward reliable and temporally consistent reconstruction.

The main contributions of this work:
\begin{itemize}
    \item We propose the unified diffusion-based framework \M for multi-temporal remote sensing cloud removal, which jointly integrates spatiotemporal modeling, region-aware optimization, and guided sampling into a coherent design.

    \item We propose a Multi-Temporal Conditional Diffusion Network (MTCDN) that explicitly models spatiotemporal dependencies across multi-temporal observations within a conditional diffusion framework, enabling effective utilization of complementary surface cues.

    \item We design a region-aware diffusion loss function that adaptively emphasizes cloud-dominated regions while preserving reliable background structures during training, encouraging the model to focus on highly uncertain regions while preserving structural integrity in reliable areas. In addition, we design a deterministic resampling strategy to refine unreliable predictions during the sampling process.

    \item Extensive experiments on two multi-temporal remote sensing datasets demonstrate that \M consistently outperforms SOTA methods, achieving improved spatial fidelity, structural preservation, and temporal consistency.
\end{itemize}

\section{Related Works}

\subsection{Learning-Based Cloud Removal Methods}

Early cloud removal methods mainly rely on mathematical and physical modeling, as well as handcrafted rules to exploit correlations among multi-temporal images and auxiliary data~\cite{tseng2008automatic, helmer2005cloud, le2009use}. However, these methods usually require strict acquisition conditions and often fail to handle complex cloud structures, resulting in limited reconstruction accuracy.

With the development of deep learning, learning-based methods have been widely adopted for cloud removal. Early neural network based approaches employ convolutional neural networks and attention mechanisms to directly learn mappings from cloud-contaminated images to cloud-free outputs. Representative methods include spatial-attention-based models~\cite{pan2020cloud}, CNN-based methods~\cite{lee2019cloud}, latent-space-oriented AEs~\cite{sintarasirikulchai2018multi}, and PMAA~\cite{zou2023pmaa}.
These approaches improve reconstruction performance compared to traditional techniques, but their point-estimation formulation limits the explicit modeling of uncertainty induced by cloud contamination.

\subsection{GAN-Based Cloud Removal Methods}

To further enhance generation quality, generative models based on Generative Adversarial Networks (GANs) have been introduced into cloud removal. Representative methods leverage adversarial learning and temporal redundancy to synthesize cloud-free imagery, including McGAN~\cite{enomoto2017filmy}, STGAN~\cite{sarukkai2020cloud}, and CTGAN~\cite{huang2022ctgan}.

Although GAN-based methods can produce visually sharper results, they often suffer from training instability and may introduce hallucinated structures in heavily cloud-covered regions. Moreover, adversarial objectives primarily emphasize perceptual realism, which may compromise radiometric consistency and physical interpretability in multi-temporal remote sensing scenarios.

\subsection{Diffusion-Based Cloud Removal}

Recently, diffusion models have been introduced for remote sensing cloud removal and shown promising results on multi-temporal imagery~\cite{jing2023denoising}. 
SeqDMs~\cite{zhao2023cloud} leverage multi-temporal cloud detection masks as auxiliary guidance within diffusion models, while DiffCR~\cite{zou2024diffcr} improves adaptability to remote sensing data through network modifications and conditional embeddings. 
Nevertheless, existing methods exploit multi-temporal information in a limited and coarse manner and largely follow conventional diffusion sampling from Gaussian noise, leaving substantial room for improvement in sampling efficiency and reconstruction quality.

To address these issues, Liu et al.~\cite{liu2025effective} proposed EMRDM, which initiates the diffusion process directly from cloudy images and formulates cloud removal as a mean-reverting diffusion task. EMRDM further employs an ordinary differential equation (ODE)-based sampler under the EDM framework\cite{karras2022elucidating} to improve sampling efficiency, and adopts a Transformer-based backbone derived from HDiT~\cite{crowson2024scalable} to fuse multi-temporal information. However, its exploitation of multi-temporal information remains limited, while our method advances diffusion-based cloud removal by more comprehensively integrating multi-temporal cues throughout the diffusion framework.

\section{Method}
\subsection{Problem Definition}
The cloud removal task aims to reconstruct a cloud-free image from cloud-contaminated satellite observations by exploiting spatial, temporal, and multimodal information.  

Let \(\mathbf{X} \in \mathbb{R}^{T \times H \times W \times C}\) denote a series of multispectral optical images captured at the same geographic location over \(T\) temporal observations, where each frame \(\mathbf{x}_t \in \mathbb{R}^{H \times W \times C}\) has spatial dimensions \((H, W)\) and \(C\) spectral channels (e.g., \(C=3\) for RGB). The sequence \(\mathbf{X} = \{\mathbf{x}_1, \mathbf{x}_2, \dots, \mathbf{x}_{T}\}\) contains varying levels of cloud coverage across time in the same location. We denote by \(\{ \mathbf{X}, \mathbf{y} \}\) a set of paired cloudy and cloud-free images, where \(\mathbf{y}\) is the corresponding cloud-free version. Optionally, a set of auxiliary messages \(\mathbf{A}\) can be provided to offer complementary structural or spectral information. 

Accordingly, the learning objectives can be formulated as conditional distributions:
\begin{equation}
P({\mathbf{y}} \mid {\mathbf{x}}_1, \dots, {\mathbf{x}}_{T}, \mathbf{A}).
\end{equation}

The final objective is to learn the conditional mapping function \( f_\theta \) that approximates these distributions and generates high-quality cloud-free reconstructions:
\begin{equation}
\hat{\mathbf{y}} = f_\theta({\mathbf{x}}_1, \dots, {\mathbf{x}}_{T}, \mathbf{A}) \approx \mathbf{y}.
\end{equation}

\subsection{Preliminaries}

In this section, we introduce the diffusion formulation adopted in this work and briefly review the mean-reverting diffusion paradigm underlying our framework.

\subsubsection{SDEs}

Score-based generative modeling via stochastic differential equations (SDEs)~\cite{song2020score} formulates diffusion as a continuous-time stochastic process that gradually perturbs clean data into noise.

Formally, given a clean target image $\mathbf{y}$, the forward (diffusion) process is modeled as a stochastic differential equation:
\begin{equation}
\dd \mathbf{x} = f(\mathbf{x}, t)\dd t + g(t) \dd\mathbf{w},
\end{equation}
where $\mathbf{x}(0) \sim p_0$ denotes the data distribution, $\mathbf{x}(T) \sim p_T$ denotes the prior (typically Gaussian), $f(\cdot, t)$ is the drift coefficient, $g(t)$ is the diffusion strength, and $\mathbf{w}$ represents a standard Wiener process.
The reverse (denoising) process can be derived as another SDE:
\begin{equation}
\dd \mathbf{x} = \left[f(\mathbf{x}, t) - g^2(t)\nabla_{\mathbf{x}}\log p_t(\mathbf{x})\right]\dd t + g(t)\dd \bar{\mathbf{w}},
\label{eq:reverse_sde}
\end{equation}
where $\nabla_{\mathbf{x}}\log p_t(\mathbf{x})$ is the score function, which can be approximated by a neural network $s_\theta(\mathbf{x}, t)$. By solving this reverse SDE, clean cloud-free reconstructions $\hat{\mathbf{y}}$ can be sampled from the noisy observations.

\subsubsection{IR-SDEs}

To adapt diffusion models for image restoration tasks such as cloud removal, IR-SDE~\cite{luo2023image} introduces a mean-reverting diffusion process that explicitly incorporates the clean target as the mean state. Specifically, the forward process is defined as
\begin{equation}
\dd \mathbf{x} = \theta_t(\boldsymbol{\mu} - \mathbf{x})\dd t + \sigma_t \dd\mathbf{w},
\end{equation}
where $\boldsymbol{\mu}$ denotes the latent target, and $\theta_t$ and $\sigma_t$ control the mean-reversion strength and noise intensity, respectively.

This formulation injects related mean information directly into the diffusion dynamics, guiding the stochastic trajectory toward the desired clean image during corruption. The resulting forward process remains a continuous-time SDE, analogous to the general diffusion formulation in Eq.~(\ref{eq:reverse_sde}), while being biased by the mean information.

\subsubsection{EMRDM}

To enhance the mean-reverting diffusion process for remote sensing cloud removal, EMRDM ~\cite{liu2025effective} modifies the forward SDE as:
\begin{equation}
\dd \mathbf{x} = f(t)(\boldsymbol{\mu} - \mathbf{x})\dd t + g(t)\dd \mathbf{w}_t,
\end{equation}
where $\boldsymbol{\mu}$ denotes the cloudy observation, and $f(t)$, $g(t)$ control the mean-reverting and diffusion strength, respectively.

To simplify the derivation, EMRDM introduces a scaled variable $\tilde{\mathbf{x}}(t) = \mathbf{x}(t) / s(t)$, leading to a Gaussian transition:
\begin{equation}
p_{0t}\big(\tilde{\mathbf{x}}(t)\mid \tilde{\mathbf{x}}_0(t)\big) = \mathcal{N}\big(\tilde{\mathbf{x}}(t); \, \tilde{\mathbf{x}}_0(t), \sigma(t)^2\mathbf{I}\big),
\end{equation}
with
\begin{equation}
\tilde{\mathbf{x}}_0(t) = \mathbf{x}(0) + \frac{1 - s(t)}{s(t)}\boldsymbol{\mu}.
\end{equation}

During denoising, the reverse ODE is expressed as (corresponding to a deterministic sampling trajectory, in contrast to stochastic SDE-based sampling):

\begin{equation}
\begin{aligned}
\dd \tilde{\mathbf{x}}(t)
&= \left[-\frac{\dot s(t)}{s(t)^{2}}\boldsymbol{\mu}\right.\\
&\quad\left.-\frac{\dot{\sigma}(t)}{\sigma(t)}
  \bigl(D_\theta(\tilde{\mathbf{x}}(t);\sigma(t),\mathbf c)
   + \tfrac{1-s(t)}{s(t)}\boldsymbol{\mu}
   - \tilde{\mathbf{x}}(t)\bigr)\right]\dd t ,
\end{aligned}
\end{equation}
where $D_\theta(\cdot)$ is a neural network that predicts the clean reconstruction, conditioned on auxiliary features $\mathbf{c}$ .

By choosing suitable $s(t)$ and $\sigma(t)$, EMRDM effectively balances noise injection and mean-reversion, enabling stable training and accurate reconstruction of cloud-free images.

\subsection{Framework Overview}

\begin{figure}[t]
    \centering
    \includegraphics[width=\linewidth]{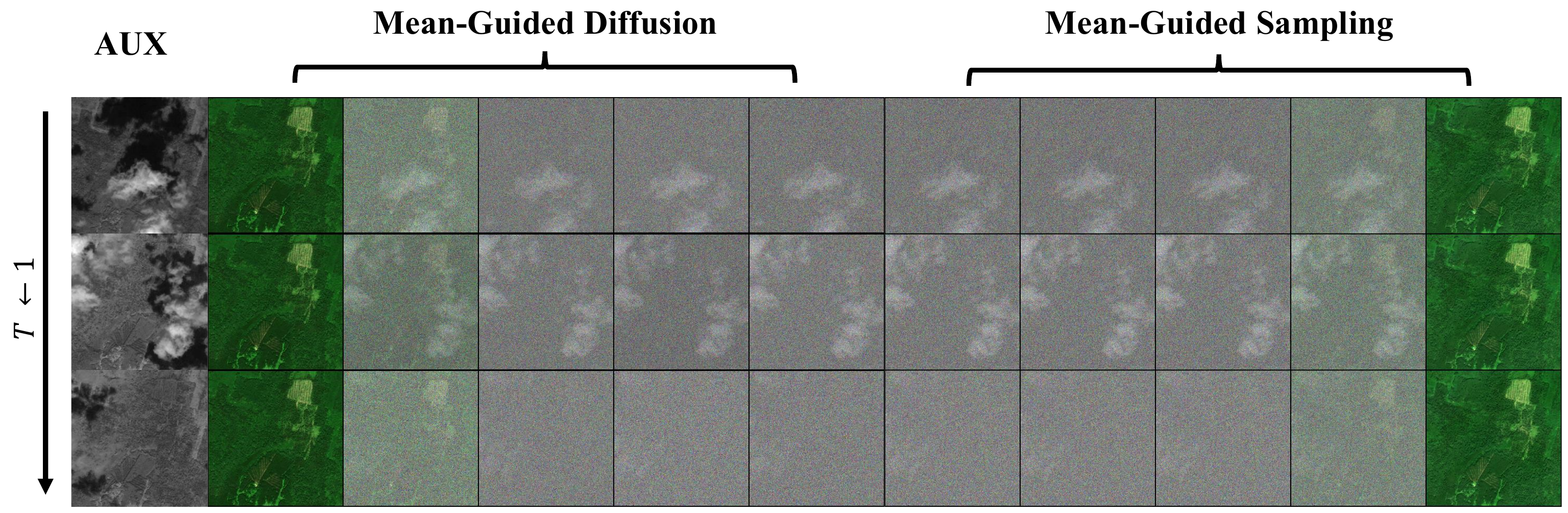}
    \caption{The diffusion workflow on multi-temporal remote sensing data.
    The first column shows the auxiliary image. Columns 2-6 correspond to the mean-guided forward diffusion process.
    Columns 7-11 illustrate the mean-guided sampling process for cloud removal. 
    }
    \label{fig:dual_channel_pipeline}
    \vspace{-2ex}
\end{figure}

\M is a multi-temporal diffusion-based cloud removal framework that integrates structural and temporal cues into both diffusion modeling and inference.
Instead of relying solely on noise-driven diffusion, the model incorporates a mean-guided diffusion formulation to provide a stable structural reference during the denoising process, enabling balanced evolution between stochastic perturbations and structural constraints ({see Fig.~\ref{fig:dual_channel_pipeline}}).
Based on this formulation, the overall design of \M consists of three complementary components: a denoising prediction network, a cloud-aware training loss, and a corrective sampling strategy.

At the network level, \M adopts a multi-temporal conditional denoising architecture that explicitly exploits temporal consistency and structural complementarity across observations.
By integrating conditional representations with temporal fusion mechanisms, the network adaptively leverages reliable surface cues from different time steps, leading to improved stability and structural fidelity in cloud removal.

During training, a cloud-aware loss is introduced to emphasize cloud-contaminated regions while preserving reliable background structures.
A mild brightness consistency constraint is further incorporated to regularize cross-temporal appearance variations caused by different acquisition conditions.

During sampling, a deterministic resampling strategy is employed to mitigate error accumulation in deterministic diffusion inference.
By selectively refining unreliable predictions under fixed diffusion steps, the proposed strategy acts as a corrective refinement mechanism, improving robustness and temporal consistency with limited additional overhead.

\subsection{Multi-Temporal Conditional Denoising Network}
\begin{figure*}[t]
    \centering
    \includegraphics[width=0.8\textwidth]{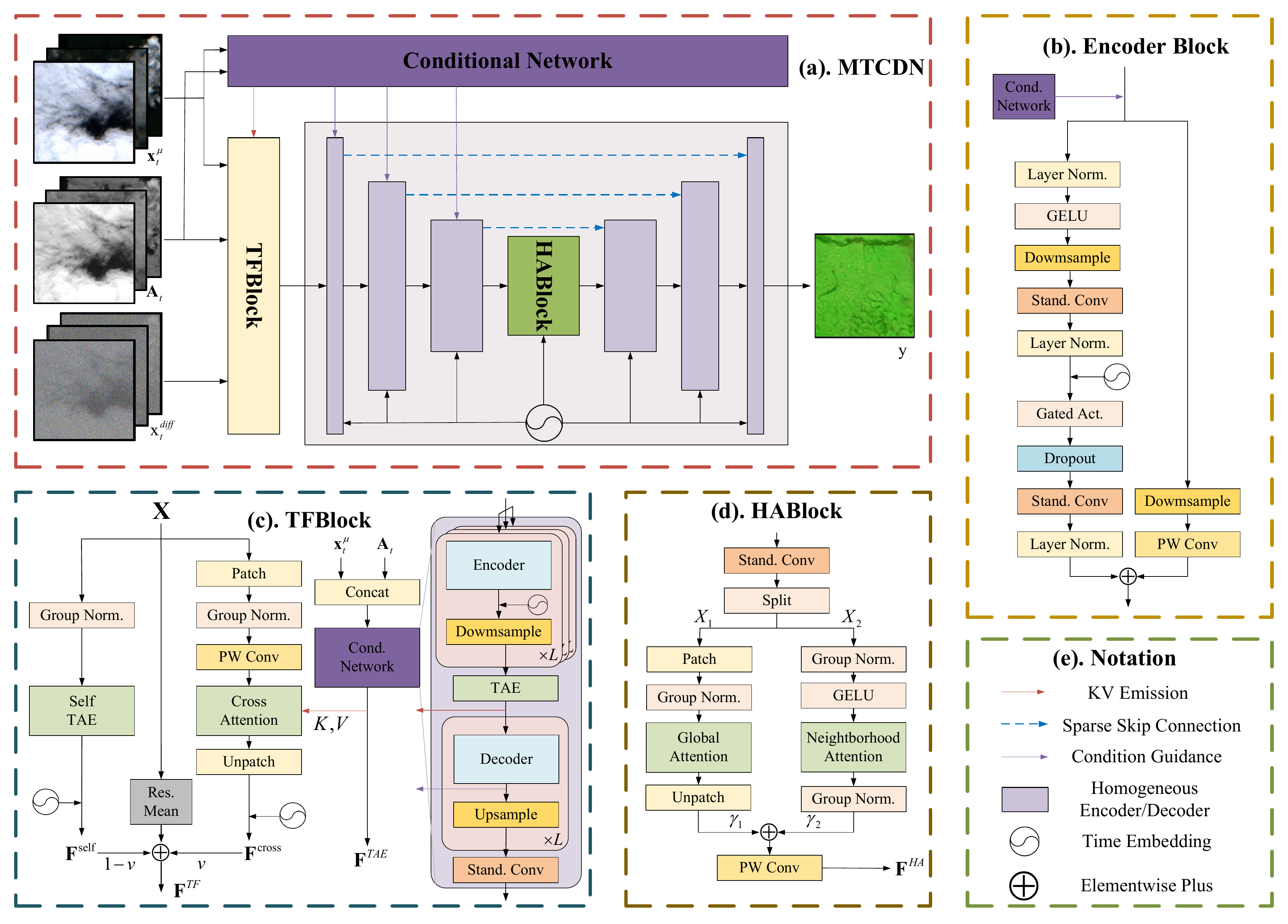}
    \caption{Architecture of the MTCDN. Key components include: (a) a U-Net-like diffusion backbone and a conditional network, where the backbone takes the diffusion target $\mathbf{x}_t^{\mathrm{diff}}$, cloudy mean image $\mathbf{x}_t^{\mu}$, and auxiliary data $\mathbf{A}_t$ as input, and  the conditional network processes $\mathbf{x}_t^{\mu}$ and $\mathbf{A}_t$; (b) encoder block with downsampling and conditional modulation; (c) Temporal Fusion Block (TFBlock) using self-attention and cross-attention for temporal feature fusion; (d) Hybrid Attention Block (HABlock) combining global and neighborhood attention; (e) notation legends. 
    }
    \label{fig:architecture}
    \vspace{-3ex}
\end{figure*}

The denoising network, termed \textbf{MTCDN} (Multi-Temporal Conditional Diffusion Network), 
is specifically designed for multi-temporal cloud removal tasks. As illustrated in Figure~\ref{fig:architecture}, it integrates several key architectural modules, including a \textbf{backbone network} for feature extraction, a \textbf{conditional network} for processing structural conditioning signals, a temporal fusion block (\textbf{TFBlock}) for modeling dependencies across temporal observations, and a hybrid attention block (\textbf{HABlock}) to enhance spatial-spectral representation learning. MTCDN is a scalable and flexible architecture that can be adapted to various remote sensing scenarios by adjusting network hyperparameters, thereby controlling the model complexity and computational cost to suit different dataset scales and task requirements.

\subsubsection{Backbone Network and Conditional Network}

The backbone network is built upon a U-Net-like diffusion architecture~\cite{ronneberger2015u} with symmetric downsampling and upsampling pathways separated by a middle block~\cite{karras2024analyzing}. The input consists of the diffusion target \({\mathbf{x}}^{diff}_t\), the cloudy mean image \({\mathbf{x}}_t^{\mu}\), and optional auxiliary information \(\mathbf{A_t}\). These items are concatenated along the channel dimension to form the complete conditional input. The conditional representation provides structural and spectral priors that are robust to cloud contamination.

The conditional network processes the conditional input (\({\mathbf{x}}^{cond}_t=\mathbf{x}_t^{\mu}\oplus \mathbf{A_t}\)) through a lightweight encoder-decoder structure to extract multi-level features. Each temporal slice is encoded independently, followed by a temporal attention that fuses information across multiple observations. The fused attention keys and values are passed to the backbone network as external guidance for cloud removal. To ensure seamless integration, the output features of the conditional network are spatially aligned with the downsampling stages of the backbone network, enabling conditional embeddings to be injected at corresponding scales for enhanced feature modulation.

The backbone network takes the diffusion input together with the conditional features, and processes them via a hierarchical feature extraction pipeline. To effectively exploit temporal information from multi-temporal observations, the backbone uses a TFBlock to adaptively fuse temporal cues. In addition, a HABlock is employed to jointly model spatial and spectral dependencies, enabling the network to focus on relevant regions and restore missing information in heavily cloud-covered areas.

Through the combination of the conditional network and the backbone network, the MTCDN achieves a balance between structural priors and dynamic temporal cues. The final output of this stage is a set of refined feature maps that contain comprehensive spatiotemporal information, which are then decoded into high-quality, cloud-free reconstructions.

\subsubsection{TFBlock}
The \textbf{Temporal Fusion Block} is designed to model temporal dependencies across multiple observations \(\mathbf{X} = \{\mathbf{x}_1, \mathbf{x}_2, \dots, \mathbf{x}_{T}\}\), enabling the network to distinguish persistent surface information from transient cloud effects. 

In the conditional network, the TAE (Temporal Attention Encoder) \cite{garnot2020satellite} is employed in the intermediate layers. Given a set of encoded temporal features \(\{\mathbf{F}_t^{cond}\}_{t=1}^{T}\), the self-attention mechanism learns adaptive relationships between different time steps, generating a fused representation:
\begin{equation}
\mathbf{F}^{TAE}, \mathbf{KV} = \text{TAE}(\{\mathbf{F}_t^{cond}\}_{t=1}^{T}),
\end{equation}
where \(\mathbf{KV}\) denotes the key and value calculated by self-attention mechanism. This module aggregates complementary information from multiple observations, producing cloud-invariant conditional features.

In the backbone network, the TFBlock is applied at the early encoding stage to jointly model temporal dependencies via a combination of temporal self-attention and temporal cross-attention. Specifically, temporal self-attention is employed to capture intrinsic correlations among multi-temporal diffusion representations, while temporal cross-attention is used to align the diffusion features with fused conditional representations derived from auxiliary temporal cues~\cite{lin2022cat}.
\begin{equation}
\mathbf{F}^{\mathit{self}} = \text{SelfAttn}(\{\mathbf{x}_t\}_{t=1}^{T}),
\end{equation}
\begin{equation}
\mathbf{F}^{\mathit{cross}} = \text{CrossAttn}(\{\mathbf{x}_t\}_{t=1}^{T}, \mathbf{KV}),
\end{equation}
\begin{equation}
\mathbf{F}^{\mathit{TF}} = v \cdot \mathbf{F}^{\mathit{cross}} + (1 - v) \cdot \mathbf{F}^{\mathit{self}},
\label{eq:adapt}
\end{equation}
Eq.~(\ref{eq:adapt}) allows the network to adaptively balance guided temporal alignment and independent temporal reasoning, enabling robust exploitation of cloud-free cues while preventing over-reliance on conditional information under noisy or uncertain observations\cite{arevalo2017gated}.

\subsubsection{HABlock}

To further enhance spatial-spectral representation learning, a \textbf{Hybrid Attention Block} is integrated into the middle stage of the backbone network. The HABlock jointly exploits \emph{global attention} and \emph{neighborhood attention} to capture both long-range dependencies and local contextual relationships within each feature map \(\mathbf{F}^{i}\), i.e.,
\begin{equation}
\mathbf{F}^{HA} = \gamma_1\text{GlobalAttn}(\mathbf{F}^{i}) + \gamma_2\text{NeighborhoodAttn}(\mathbf{F}^{i}),
\end{equation}
where \(\text{GlobalAttn}(\cdot)\) models full-image dependencies to recover large-scale structures and eliminate cloud shadows~\cite{vaswani2017attention}, while \(\text{NeighborhoodAttn}(\cdot)\) focuses on local consistency, refining fine-grained spatial textures and spectral details~\cite{hassani2023neighborhood}. \(\gamma_1\) and \(\gamma_2\) are generated according to timestamps. The combination of these two complementary attention mechanisms allows the network to simultaneously maintain global coherence and local fidelity in reconstructed cloud-free images.

\subsection{Cloud-aware Loss with Brightness Consistency}

\begin{figure}[t]
    \centering
    \includegraphics[width=0.48\textwidth]{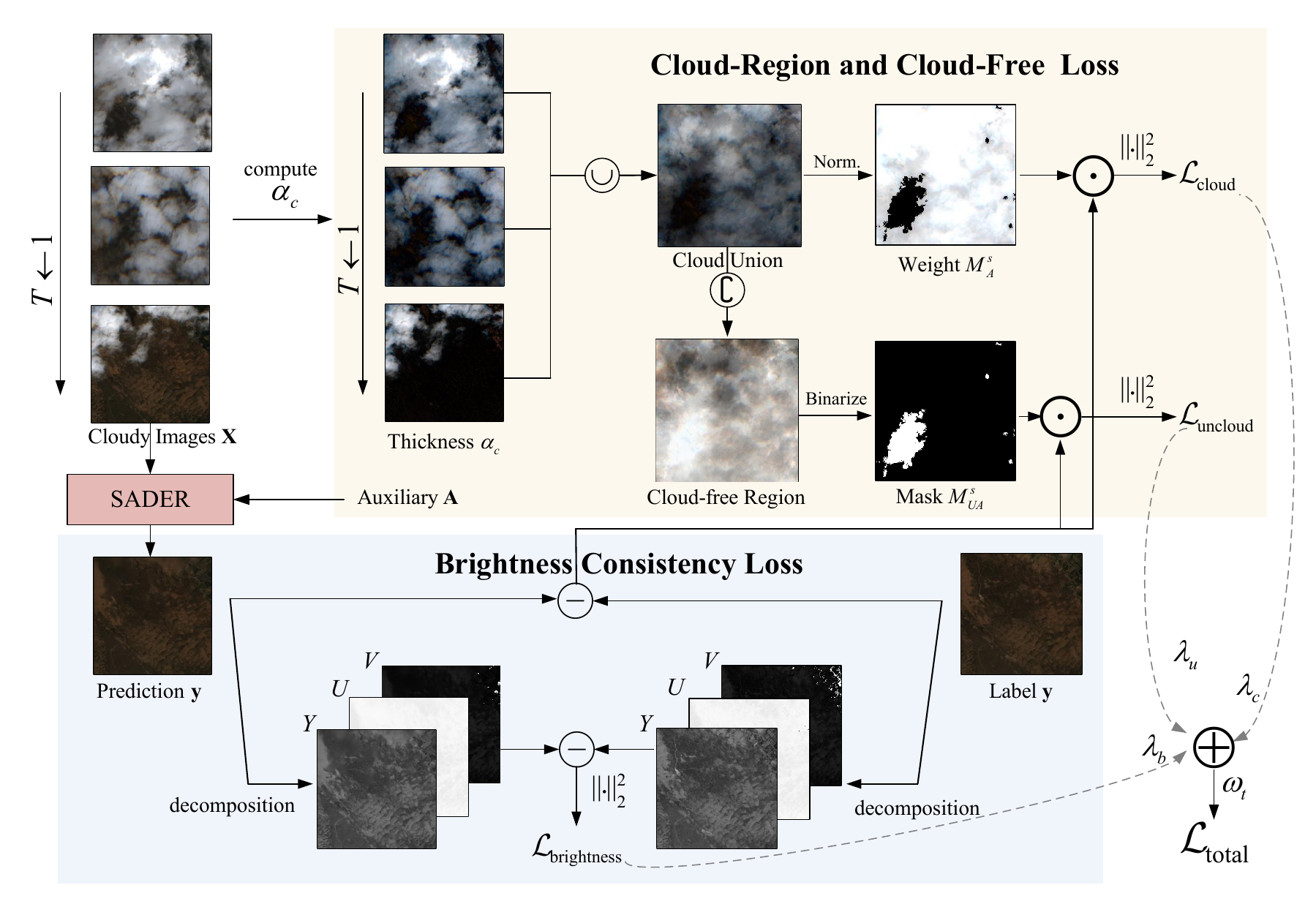}
    \caption{Pipeline of the cloud-aware attention loss.
    Multi-temporal cloudy observations are first used to construct region-aware masks for separating cloud and cloud-free areas. Here, $\cup$ denotes the temporal union of multi-temporal cloudy images to form the cloud union image. $\complement$ denotes its complement to obtain the cloud-free region. $\mathrm{Norm.}$ normalizes cloud thickness to $[0,1]$ to generate thickness-aware weights, and $\mathrm{Binarize}$ produces a binary mask. 
    }
    \label{fig:cloud_aware_loss}
    \vspace{-3ex}
\end{figure}

To further enhance the denoising effectiveness of the diffusion model in cloud-contaminated regions, we introduce a cloud-aware attention loss in the \M framework. 
As shown in Fig.~\ref{fig:cloud_aware_loss}, the overall loss integrates multiple components that focus on different spatial regions and spectral constraints, while incorporating a time-dependent weight \(w_t\) that follows the noise scheduling principle of EMRDM~\cite{liu2025effective}.
The total loss is formulated as:
\begin{equation}
\mathcal{L}_{\text{total}} 
= w_t \cdot \left(
\lambda_c \cdot \mathcal{L}_{\text{cloud}}
+ \lambda_u \cdot \mathcal{L}_{\text{uncloud}}
+ \lambda_b \cdot \mathcal{L}_{\text{brightness}}
\right),
\label{eq:arsedm_total_loss}
\end{equation}
where \( \lambda_c \), \( \lambda_u \), and \( \lambda_b \) are hyperparameters that balance the contributions of each component. The three terms respectively correspond to the cloud-region reconstruction, unclouded-region preservation, and brightness consistency constraints.

\subsubsection{Cloud-Region Loss}
To enhance reconstruction under heavy cloud occlusion, we introduce a \emph{cloud weight map} \(M_A^s \in [0,1]^{H \times W}\) that assigns higher weights to pixels with thicker clouds. The map is derived from the difference between the mean cloudy image \(\mathbf{x}^\mu\) and the cloud-free image \(\mathbf{y}\):
\[
V_d = \left| \mathbf{y} - \mathbf{x}^\mu \right|.
\]
Since only cloudy regions contribute to significant intensity differences, \(V_d\) effectively indicates potential cloud coverage.  
To further estimate the cloud thickness (or transparency), we adopt the following physical approximation~\cite{rozanov2004semianalytical}:
\begin{equation}
I_{\text{union}} = (1 - \alpha_c) I_{\text{ground}} + \alpha_c I_{\text{cloud}},
\quad
\alpha_c = \frac{V_d}{I_{\text{cloud}} - I_{\text{ground}}},
\label{eq:cloud_alpha}
\end{equation}
where \(I_{\text{union}}\) is the observed cloudy image, \(I_{\text{ground}}\) is the true ground reflectance (i.e., \(\mathbf{y}\)), and \(I_{\text{cloud}}\) represents the cloud radiance (typically set to 1.0 after normalization).  
The transparency map \(\alpha_c\) is normalized spatially and averaged along the temporal dimension to form the initial cloud weight map \(M_A^s\). 
The cloud-region loss is expressed as:
\begin{equation}
\mathcal{L}_{\text{cloud}} 
= \mathbb{E} \left[
M_A^s \odot \left| \hat{\mathbf{y}} - \mathbf{y} \right|^2
\right],
\end{equation}
which directs optimization toward denser cloud areas, improving reconstruction accuracy where occlusion is severe.

\subsubsection{Cloud-Free Loss}
Although diffusion noise is injected uniformly across the entire image, it may introduce undesired perturbations even in cloud-free regions\cite{croitoru2023diffusion}. To preserve the integrity of these areas, we apply a binary mask $M_{UA}^s \in \{0,1\}^{H \times W}$, yielding:
\begin{equation}
\mathcal{L}_{\text{uncloud}} 
= \mathbb{E} \left[
M_{UA}^s \odot \left| \hat{\mathbf{y}} - \mathbf{y} \right|^2
\right].
\end{equation}
This term ensures accurate reconstruction of cloud-free pixels, maintaining spatial coherence and preventing artifacts caused by denoising dynamics.

\subsubsection{Brightness Consistency Loss}

In multi-temporal datasets, variations in solar illumination, atmospheric conditions, and acquisition angles often lead to significant brightness discrepancies among observations, even for cloud-free pixels~\cite{zhu2018automatic}. 
To enforce photometric consistency while preventing undesired color distortion, we introduce a brightness-dominant consistency constraint in the YUV color space:

\begin{equation}
\mathcal{L}_{\text{brightness}}
= \mathbb{E} \left[
\left\|
\text{YUV}(\hat{\mathbf{y}}) - \text{YUV}(\mathbf{y})
\right\|_2^2
\right],
\label{eq:brightness_yuv}
\end{equation}
where $\text{YUV}(\cdot)$ denotes the RGB-to-YUV color space transformation~\cite{podpora2014yuv}. 
Although the loss is applied uniformly across all three channels, the luminance component $Y$ inherently dominates the YUV representation. As a perceptually weighted combination of RGB that models human brightness sensitivity, $Y$ carries the majority of image energy and structural detail, whereas the chrominance channels $U$ and $V$, being zero-mean color difference signals, exhibit much smaller magnitudes and contribute primarily to hue rather than illumination~\cite{poynton2012digital}. Consequently, the optimization is naturally biased toward brightness consistency, with $U$ and $V$ acting as mild regularizers to suppress spurious color shifts.

Since the RGB-to-YUV transformation is linear, this loss can be seamlessly integrated into the diffusion objective without affecting the theoretical weighting scheme. 
Overall, the proposed loss in Eq.~\ref{eq:arsedm_total_loss} encourages physically plausible brightness recovery while maintaining stable color consistency across multi-temporal observations.

\subsection{Deterministic Resampling Strategy}

To mitigate cumulative errors arising from deterministic diffusion sampling~\cite{song2020denoising} and enhance reconstruction fidelity~\cite{karras2022elucidating}, we propose a \textbf{deterministic resampling strategy} within the \M framework.
Rather than relying on a single deterministic trajectory, the proposed strategy performs repeated perturbation at the same diffusion level by reintroducing noise and mean components, followed by denoising operations to generate multiple candidate predictions.
A corrective refinement is then applied through adaptive fusion guided by structural priors extracted from a pretrained Masked Autoencoder (\textit{MAE})~\cite{he2022masked}, thus improving sampling stability and reconstruction accuracy.

\subsubsection{MAE-Guided Structural Prior}

To provide structure-aware guidance during deterministic resampling, we introduce a plug-and-play structural prior extractor that operates independently of the diffusion model.
Specifically, we adopt a pretrained \textit{MAE}~\cite{he2022masked} implemented with a ViT-style asymmetric encoder--decoder architecture~\cite{dehghani2023scaling}.
The MAE is pretrained on the EuroSAT-MS dataset~\cite{helber2019eurosat}, which contains approximately 27{,}000 multi-spectral remote sensing images.
Through masked reconstruction, the MAE learns global spatial layouts and low-frequency structural patterns, making it suitable as a structural reference for resampling.
After pretraining, the MAE is frozen and used solely at the sampling stage.

\subsubsection{Resampling Procedure}

At each diffusion step, two or more candidate predictions are obtained by repeatedly reintroducing noise and mean components~\cite{liu2024residual}.
A pixel-wise comparison is then conducted between the candidates with respect to the MAE-derived structural prior.
If the deviation of a prediction from the MAE prior exceeds a predefined threshold (set according to the average cloud coverage ratio), the corresponding pixel is adaptively replaced by its counterpart from the alternative prediction.
This selective fusion corrects unreliable regions while preserving overall consistency with the diffusion trajectory and mean guidance~\cite{lugmayr2022repaint}.

\subsubsection{Algorithmic Workflow}

The complete deterministic resampling procedure is summarized in Algorithm~\ref{algo:STRIA_cr}.
The strategy is seamlessly integrated into the sampling pipeline of \M and applied at each diffusion step.
Through repeated perturbation, MAE-guided evaluation, and pixel-wise fusion, the proposed deterministic resampling strategy effectively suppresses accumulated sampling errors, leading to reconstructions that are both structurally consistent and visually faithful.
All variables indexed by \(t\) denote synchronous multi-temporal data processed jointly across temporal slices.

\begin{algorithm}[ht]
  \small
  \caption{SADER-DR: Deterministic Resampling Process}
  \label{algo:STRIA_cr}
  \begin{algorithmic}[1]
    \Require Denoising network $D_{\theta}(\cdot)$; \textit{MAE} guide $MAE(\cdot)$; mean images $x_t^{\mu}$; auxiliary information $c$; sample steps $N$; resample steps $N^r$; mean response rate $\alpha$; threshold $T_h$
    \Ensure Resampled result $x^N$
    \State Sample $x_t^0 \sim \mathcal{N}(\alpha \sigma_0 x_t^{\mu}, \sigma_0^2 I)$
    \For{$i = 0$ to $N-1$}
      \State $x_t^f \gets D_{\theta}(x_t^i; \sigma_i; c), \quad x_t^{f,0} \gets x_t^f$
      \For{$j = 1$ to $N^r$}
        \State Sample $\epsilon_t^i \sim \mathcal{N}(0, \sigma_i^2 I)$
        \State $x_t^{i,\mu} \gets x_t^f + \alpha \sigma_i x_t^{\mu}$, \quad $x_t^{i,d} \gets x_t^{i,\mu} + \epsilon_t^i$
        \State $x_t^{f,j} \gets D_{\theta}(x_t^{i,d}; \sigma_i; c)$, \quad $G_j \gets MAE(x_t^{i\mu})$
        \For{pixels where $|x_t^{f,(j-1)} - x_t^{\mu}| \in \mathrm{Top}_{T_h}$}
          \State $m \gets \text{Compare } |x_t^{f,(j-1)} - G_j| \text{ and } |x_t^{f,j} - G_j|$
        \EndFor
        \State $x_t^f \gets m \odot x_t^{f,j} + (1 - m) \odot x_t^{f,(j-1)}$
      \EndFor
      \State $d_t^i \gets \frac{x_t^{i,d} - x_t^f}{\sigma_i}$
      \State $x_t^{i+1} \gets x_t^{i,d} + (\sigma_{i+1} - \sigma_i) d_t^i$
    \EndFor
    \State $x^N \gets \text{mean}_T(x_t^N)$
    \State \Return $x^N$
  \end{algorithmic}
\end{algorithm}

In Algorithm~\ref{algo:STRIA_cr}, $\sigma_i$ controls the noise intensity at step $i$, and $\alpha$ determines the mean-reversion rate toward the cloudy mean response.
At each step, $x_t^{k,\mu}$ denotes the intermediate sample after injecting the step-wise mean information, and $x_t^{k,d}$ represents the sample obtained by further adding Gaussian noise.
The variable $x_t^{f,k}$ corresponds to the denoised output produced by the network at the current iteration, where the index $k$ may refer to $i$, $j$, or be omitted depending on the specific sampling stage.
Corrective process is applied to pixels whose deviation $|x_t^{f,(j-1)} - x_t^{\mu}|$ lies in the top $T_h$ fraction among all pixels, indicating the most unstable predictions to be selectively refined.

\section{Experiments}

\subsection{Baseline Methods}
We compare the proposed method with representative multi-temporal remote sensing cloud removal approaches spanning three major categories.

\textbf{Learning-based cloud removal methods}: Pix2Pix~\cite{isola2017image}, AE~\cite{sintarasirikulchai2018multi}, PMAA~\cite{zou2023pmaa}, CR-TS Net~\cite{ebel2022sen12ms}, and UnCRtainTS~\cite{ebel2023uncrtaints}. 
These methods typically learn deterministic mappings from cloud-contaminated images to cloud-free outputs.

\textbf{GAN-based cloud removal methods}: McGAN~\cite{enomoto2017filmy}, CycleGAN~\cite{isola2017image}, STGAN~\cite{sarukkai2020cloud}, and CTGAN~\cite{huang2022ctgan}.
These methods leverage adversarial learning to enhance visual quality and exploit temporal redundancy. 

\textbf{Diffusion-based generative models}: DDPM-CR~\cite{jing2023denoising}, SeqDMs~\cite{zhao2023cloud}, DiffCR~\cite{zou2024diffcr}, and EMRDM~\cite{liu2025effective}. These methods formulate cloud removal as a progressive denoising process. Specifically, EMRDM introduces a mean-reverting diffusion formulation by guiding the denoising process toward a cloudy mean response, serving as a strong diffusion-based baseline.

\subsection{Datasets}

We evaluate the proposed method on two widely used multi-temporal cloud removal benchmarks: 

\textbf{Sen2\_MTC\_New} dataset consists of multi-temporal Sentinel-2 optical images paired with SAR observations used as auxiliary conditional information. The reconstruction target is the cloud-free RGB optical image.
Raw optical pixel values are divided by 10000 and rescaled to $[0,1]$, followed by normalization with mean $0.5$ and standard deviation $0.5$.
During diffusion training, all inputs are mapped to $[-1,1]$, and predictions are remapped to $[0,1]$ for evaluation.

\textbf{SEN12MS-CR-TS(EA)} is the Eastern Asia subset of the SEN12MS-CR dataset, which is a challenging multispectral cloud removal task.
In this setting, all 13 Sentinel-2 spectral bands are jointly reconstructed from multi-temporal cloudy observations. Pixel values are clipped to $[0,10000]$ and linearly rescaled to $[-1,1]$ during diffusion training and restored to $[0,1]$ for evaluation.

\subsection{Experimental Setup}

\textbf{Experiment environment.} 
All models are implemented using Python 3.12.11 and PyTorch 2.5.1+cu121, and trained using the Muon optimizer~\cite{jordan2024muon} on an NVIDIA A800 GPU.

\textbf{Evaluation metrics.} We use six commonly adopted metrics, including peak signal-to-noise ratio (PSNR), structural similarity index measure (SSIM)~\cite{wang2004image}, mean absolute error (MAE), root mean squared error (RMSE), spectral angle mapper (SAM)~\cite{kruse1993spectral}, and learned perceptual image patch similarity (LPIPS)~\cite{zhang2018unreasonable}. PSNR and SSIM are higher-is-better metrics, whereas the others are lower-is-better. Since LPIPS is defined on RGB images and evaluates perceptual similarity based on three-channel visual features, we use SAM for the SEN12MS-CR-TS(EA) dataset with 13 spectral bands to provide a more appropriate measure of spectral fidelity and better reflect reconstruction quality in the multi-spectral setting.

\subsection{Main Results and Analysis}
On Sen2\_MTC\_New (Table~\ref{tab:sen12ms_comparison}), GAN-based methods and autoencoder-based methods show inferior SSIM and LPIPS performance, indicating limited robustness in preserving perceptual quality and structural consistency under cloud contamination.
Diffusion-based approaches consistently outperform these methods, demonstrating stronger capability in handling uncertainty.
Among the baselines, EMRDM provides a competitive reference by incorporating conditional guidance into the diffusion process.
Built upon this baseline, \M achieves improvements of $3.41\%$, $3.85\%$, and $6.56\%$ in PSNR, SSIM, and LPIPS, respectively, indicating more accurate reconstruction across both cloud-covered and cloud-free regions.

\begin{table}[h]
\centering
\setlength{\tabcolsep}{8pt}
\begin{tabular}{l c c c}
\toprule
\textbf{Method} & \textbf{PSNR$\uparrow$} & \textbf{SSIM$\uparrow$} & \textbf{LPIPS$\downarrow$} \\
\midrule
McGAN\cite{enomoto2017filmy}              & 17.448 & 0.513 & 0.447 \\
Pix2Pix\cite{isola2017image}              & 16.985 & 0.455 & 0.535 \\
AE\cite{sintarasirikulchai2018multi}      & 15.100 & 0.441 & 0.602 \\
CycleGAN\cite{chu2017cyclegan}             & 17.678 & 0.615 & 0.392 \\
STGAN\cite{sarukkai2020cloud}              & 18.152 & 0.587 & 0.513 \\
CTGAN\cite{huang2022ctgan}                 & 18.308 & 0.609 & 0.384 \\
PMAA\cite{zou2023pmaa}                     & 18.009 & 0.614 & 0.392 \\
CR-TS Net\cite{ebel2022sen12ms}            & 18.585 & 0.615 & 0.342 \\
UnCRtainTS\cite{ebel2023uncrtaints}        & 18.770 & 0.631 & 0.333 \\
DDPM-CR\cite{jing2023denoising}            & 18.742 & 0.614 & 0.329 \\
DiffCR\cite{zou2024diffcr}                 & 19.150 & 0.531 & 0.255 \\
EMRDM\cite{liu2025effective}               & \underline{\textit{20.249}} & \underline{\textit{0.702}} & \underline{\textit{0.244}} \\
\cmidrule(lr){1-4}
\textbf{Ours (\M)}                         & \textbf{20.941} & \textbf{0.729} & \textbf{0.228} \\
Improvement                                & \textcolor[rgb]{0.55,0,0}{3.41\%$\uparrow$} & \textcolor[rgb]{0.55,0,0}{3.85\%$\uparrow$} & \textcolor[rgb]{0.55,0,0}{6.56\%$\uparrow$} \\
\bottomrule
\end{tabular}
\caption{Quantitative comparison of different cloud removal methods on the Sen2\_MTC\_New dataset. Best results are shown in bold, and second-best results are shown in italics with underlines.}
\label{tab:sen12ms_comparison}
\vspace{-2ex}
\end{table}

\begin{table}[h]
\centering
\setlength{\tabcolsep}{8pt}
\begin{tabular}{l c c c}
\toprule
\textbf{Method} & \textbf{PSNR$\uparrow$} & \textbf{SSIM$\uparrow$} & \textbf{SAM$\downarrow$} \\
\midrule
McGAN\cite{enomoto2017filmy}              & 25.279 & 0.819 & 10.051 \\
Pix2Pix\cite{isola2017image}            & 23.303 & 0.818 & 10.932 \\
CycleGAN\cite{chu2017cyclegan}           & 26.467 & 0.840 & 12.483 \\
STGAN\cite{sarukkai2020cloud}              & 25.420 & 0.818 & 12.548 \\
PMAA\cite{zou2023pmaa}               & 28.396 & 0.771 & 16.776 \\
CR-TS Net\cite{ebel2022sen12ms}  & 26.681 & 0.836 & 10.663 \\
UnCRtainTS\cite{ebel2023uncrtaints}         & {28.756} & {0.914} & {8.428} \\
DiffCR\cite{zou2024diffcr}             & 26.072 & 0.522 & 11.662 \\
SeqDMs\cite{zhao2023cloud}             & 28.074 & 0.827 & 12.777 \\
EMRDM\cite{liu2025effective}              & \underline{\textit{29.320}} & \underline{\textit{0.926}} & \underline{\textbf{}\textit{5.933}} \\
\cmidrule(lr){1-4}
\textbf{Ours (\M)} & \textbf{31.234} & \textbf{0.937} & \textbf{5.885} \\
Improvement & \textcolor[rgb]{0.55,0,0}{6.53\%$\uparrow$} & \textcolor[rgb]{0.55,0,0}{1.19\%$\uparrow$} & \textcolor[rgb]{0.55,0,0}{0.81\%$\uparrow$}\\
\bottomrule
\end{tabular}
\caption{Quantitative comparison of different cloud removal methods on the SEN12MS-CR-TS(EA). Best results are shown in bold, and the second-best results are shown in italics with underlines.}
\label{tab:sen12ms_psnr_ssim_sam}
\vspace{-2ex}
\end{table}

On the more challenging SEN12MS-CR-TS(EA) dataset (Table~\ref{tab:sen12ms_psnr_ssim_sam}), deterministic image-to-image translation methods suffer from large SAM errors, revealing pronounced spectral distortions.
While multi-temporal and diffusion-based methods alleviate this issue to some extent, EMRDM again achieves the strongest baseline performance.
The proposed \M further improves PSNR and SSIM by $6.53\%$ and $1.19\%$, respectively, while reducing SAM by $0.81\%$, demonstrating improved spectral fidelity and spatial consistency.

Overall, the consistent gains across both datasets confirm that MTCDN, cloud-aware loss, and deterministic resampling jointly enable more effective exploitation of multi-temporal information within the diffusion framework, leading to improved reconstruction quality under diverse cloud conditions.

\subsection{Visualization Analysis}

\begin{figure*}[t]
    \centering
    \includegraphics[width=0.90\textwidth]{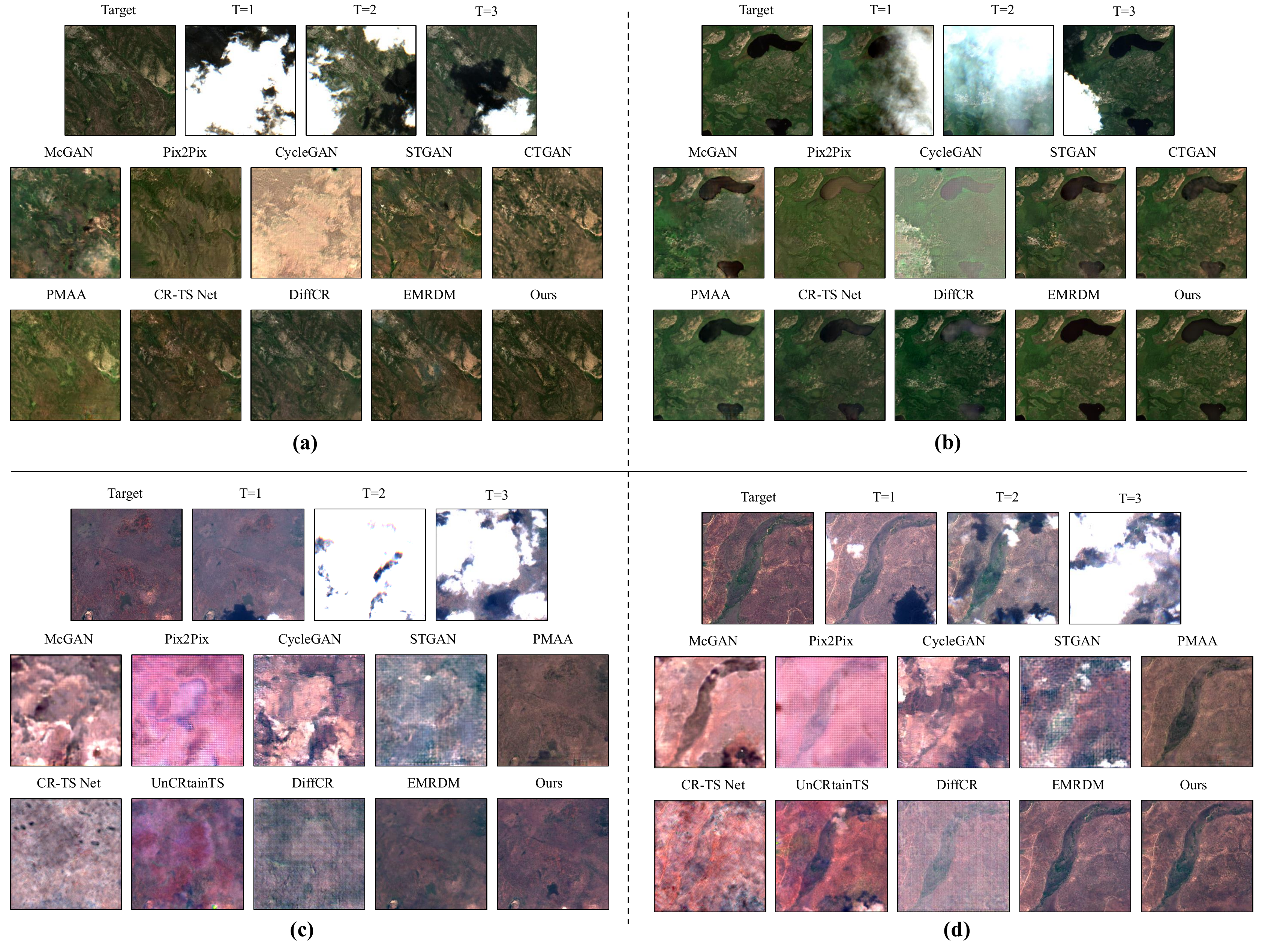}
    \caption{Visual comparison of the proposed method against representative baseline approaches on two multi-temporal cloud removal benchmarks.
    For each example, the cloud-free target and multi-temporal cloudy observations ($T\!-\!1$, $T\!-\!2$, $T\!-\!3$) are shown, followed by the corresponding reconstructions produced by different methods.
    (a)-(b) are from the Sen2\_MTC\_New dataset, while (c)-(d) are from the SEN12MS-CR-TS(EA) dataset.
    All results are visualized using the same post-processing and value rescaling protocol as EMRDM~\cite{liu2025effective}. 
    }
    \label{fig:qual_all_results}
    \vspace{-1ex}
\end{figure*}

Fig.~\ref{fig:qual_all_results} shows qualitative comparisons between \M~and baseline methods.
The proposed approach produces visually cleaner reconstructions with fewer artifacts and better structural consistency, and more closely matches the cloud-free targets.
This demonstrates the advantage of incorporating multi-temporal information into the diffusion process for robust cloud removal.

\subsection{Ablation Studies}

To verify the effectiveness of each component in \M model, we conduct three groups of ablation studies: 

\subsubsection{MTCDN Components}

\begin{table*}[t]
\centering
\setlength{\tabcolsep}{5pt}
\renewcommand{\arraystretch}{1.2}
\begin{tabular}{c c c c c c c c c c c}
\toprule
\multirow{2}{*}{\textbf{Architecture}} 
& \multicolumn{5}{c}{\textbf{Sen2\_MTC\_New}} 
& \multicolumn{5}{c}{\textbf{SEN12MS-CR-TS(EA)}} \\
\cmidrule(lr){2-6} \cmidrule(lr){7-11}
& PSNR$\uparrow$ 
& RMSE$\downarrow$ 
& MAE$\downarrow$ 
& SSIM$\uparrow$
& LPIPS$\downarrow$ 
& PSNR$\uparrow$ 
& RMSE$\downarrow$ 
& MAE$\downarrow$ 
& SSIM$\uparrow$ 
& SAM$\downarrow$ \\
\midrule
w/o TFBlock  
& 20.487 & 0.106 & 0.079 & 0.717 & 0.235
& 28.880 & 0.037 & 0.079 & 0.909 & 7.364 \\
w/o HABlock 
& 20.659 & 0.107 & 0.079 & 0.721 & 0.232
& 29.253 & 0.035 & 0.030 & 0.921 & 6.685 \\
MTCDN       
& \textbf{20.806} & \textbf{0.104} & \textbf{0.077} & \textbf{0.722} & \textbf{0.229}
& \textbf{29.914} & \textbf{0.033} & \textbf{0.028} & \textbf{0.929} & \textbf{5.973} \\
\bottomrule
\end{tabular}
\caption{Ablation study of the MTCDN architecture. We sample 5 steps without resampling.
}
\label{tab:mtcdn_ablation}
\vspace{-2ex}
\end{table*}

For MTCDN, we conduct ablation studies by individually removing TFBlock and HABlock.
According to Table~\ref{tab:mtcdn_ablation}, the complete MTCDN consistently achieves the best performance across all evaluation metrics. Removing TFBlock leads to notable performance degradation, indicating that explicit temporal fusion is critical for capturing cross-temporal consistency and stable reconstruction under cloud contamination. Removing HABlock also degrades performance, but to a relatively smaller extent, suggesting that the hybrid attention mechanism mainly contributes to refining spatial--spectral representations and local structural details.
Overall, TFBlock provides robust multi-temporal priors through temporal dependency modeling, while HABlock complements this process by enhancing structural refinement, and their joint integration enables MTCDN to achieve superior reconstruction quality.

\subsubsection{Loss Function Components}

\begin{table*}[t]
\centering
\setlength{\tabcolsep}{5pt}
\renewcommand{\arraystretch}{1.2}
\begin{tabular}{c c c c c c c c c c c}
\toprule
\multirow{2}{*}{\textbf{Loss Setting}} 
& \multicolumn{5}{c}{\textbf{Sen2\_MTC\_New}} 
& \multicolumn{5}{c}{\textbf{SEN12MS-CR-TS(EA)}} \\
\cmidrule(lr){2-6} \cmidrule(lr){7-11}
& PSNR$\uparrow$ 
& RMSE$\downarrow$ 
& MAE$\downarrow$ 
& SSIM$\uparrow$
& LPIPS$\downarrow$ 
& PSNR$\uparrow$ 
& RMSE$\downarrow$ 
& MAE$\downarrow$ 
& SSIM$\uparrow$ 
& SAM$\downarrow$ \\
\midrule
Original Loss (MSE) 
& 20.423 & 0.108 & 0.081 & 0.711 & 0.236
& 29.401 & 0.034 & 0.030 & 0.918 & 7.168 \\

w/o Cloud-Free Loss 
& 19.978 & 0.113 & 0.085 & 0.712 & 0.249
& 28.683 & 0.038 & 0.032 & 0.915 & 7.677 \\

w/o Brightness Consistency Loss 
& 20.571 & 0.107 & 0.079 & 0.721 & 0.232
& 29.693 & 0.035 & 0.029 & 0.927 & 6.162 \\

Cloud-Aware Loss 
& \textbf{20.806} & \textbf{0.104} & \textbf{0.077} & \textbf{0.722} & \textbf{0.229}
& \textbf{29.914} & \textbf{0.033} & \textbf{0.028} & \textbf{0.929} & \textbf{5.973} \\
\bottomrule
\end{tabular}
\caption{Ablation study on loss function components. 
Sampling 5 steps without resampling.}
\label{tab:loss_ablation}
\end{table*}

To evaluate the effectiveness of the proposed loss design, we perform ablation studies by selectively removing individual loss terms.
As shown in Table~\ref{tab:loss_ablation}, excluding any component leads to a clear degradation in overall performance on both datasets, indicating that the different loss terms provide complementary supervision signals.
Removing the cloud-free region constraint weakens reconstruction stability in cloud-free regions with reliable observations, while eliminating the brightness consistency term also degrades performance, highlighting the importance of cross-temporal photometric consistency.
In contrast, the full cloud-aware loss achieves the most balanced performance across datasets, suggesting that jointly constraining cloud-covered regions, preserving cloud-free areas, and regularizing brightness variations yields a more effective training objective.
In addition, we conducted a supplementary sensitivity analysis on different loss weight configurations and adopt $\lambda_c : \lambda_u : \lambda_b = 3 : 1 : 2$ as the default setting.
This choice provides a well-balanced trade-off among the different constraints.

\subsubsection{Deterministic Resampling Components}

\begin{table*}[t]
\centering
\setlength{\tabcolsep}{5pt}
\begin{tabular}{c ccccc ccccc}
\toprule
\multirow{2}{*}{$N + N^r$} 
& \multicolumn{5}{c}{\textbf{Sen2\_MTC\_New}} 
& \multicolumn{5}{c}{\textbf{SEN12MS-CR-TS(EA)}} \\
\cmidrule(lr){2-6} \cmidrule(lr){7-11}
& PSNR$\uparrow$ 
& RMSE$\downarrow$ 
& MAE$\downarrow$ 
& SSIM$\uparrow$
& LPIPS$\downarrow$
& PSNR$\uparrow$ 
& RMSE$\downarrow$ 
& MAE$\downarrow$ 
& SSIM$\uparrow$ 
& SAM$\downarrow$ \\
\midrule
$5+0$ 
& 20.806 & 0.104 & 0.077 & 0.722 & 0.229
& 29.914 & 0.033 & 0.028 & 0.929 & 5.973 \\

$5+1$ 
& 20.859 & 0.103 & 0.077 & 0.725 & 0.229
& \textbf{31.234} & \textbf{0.028} & \textbf{0.023} & \textbf{0.937} & \textbf{5.885} \\

$5+2$ 
& 20.873 & 0.103 & 0.076 & 0.726 & 0.229
& 31.256 & 0.028 & 0.023 & 0.937 & 5.884 \\

$4+0$ 
& 20.900 & 0.103 & 0.076 & 0.727 & 0.229
& 29.491 & 0.036 & 0.031 & 0.924 & 5.992 \\

$4+1$ 
& \textbf{20.941} & \textbf{0.102} & \textbf{0.076} & \textbf{0.729} & \textbf{0.228}
& 29.520 & 0.034 & 0.029 & 0.928 & 5.982 \\

$4+2$ 
& 20.972 & 0.102 & 0.075 & 0.729 & 0.227
& 29.550 & 0.034 & 0.029 & 0.928 & 5.979 \\
\bottomrule
\end{tabular}
\caption{Sensitivity analysis of sampling steps $N$ and deterministic resampling steps $N^r$.
The highlighted configurations ($4+1$ for Sen2\_MTC\_New and $5+1$ for SEN12MS-CR-TS(EA)) are selected as the final settings in our framework.}
\label{tab:sampling_resampling_final}
\end{table*}

We conduct ablation studies to analyze the effect of the proposed deterministic resampling strategy in diffusion sampling.
As shown in Table~\ref{tab:sampling_resampling_final}, introducing resampling consistently improves reconstruction performance over baseline deterministic sampling on both datasets, indicating its effectiveness in refining unstable predictions in cloud-contaminated regions.

With the number of diffusion steps $N$ fixed, increasing the number of resampling steps $N^r$ generally leads to performance improvements with diminishing returns.
Moreover, resampling cannot fully compensate for performance differences caused by insufficient diffusion steps, confirming that deterministic resampling mainly acts as a local refinement mechanism rather than a substitute for adequate diffusion depth.
Considering the trade-off between accuracy and computational cost, we adopt a single resampling step ($N^r = 1$) as the default configuration.

Table~\ref{tab:resampling_strategy_ablation} compares different corrective strategies under fixed configurations. Performance remains limited without resampling or corrective guidance, as deterministic sampling lacks robustness to accumulated errors. Guidance during resampling consistently improves quality: mean-guided correction offers moderate gains, convolution-based guidance incorporates spatial cues for further improvement, and the proposed MAE-guided strategy achieves the most consistent gains across all metrics by leveraging a frozen external prior to reliably correct unstable predictions in a plug-and-play manner. 

\begin{table*}[t]
\centering
\setlength{\tabcolsep}{6pt}
\renewcommand{\arraystretch}{1.15}
\begin{tabular}{ccccccccccc}
\toprule
\multirow{2}{*}{\textbf{Correction Strategy}} 
& \multicolumn{5}{c}{\textbf{Sen2\_MTC\_New} ($N{+}N^r{=}4{+}1$)} 
& \multicolumn{5}{c}{\textbf{SEN12MS-CR-TS(EA)} ($N{+}N^r{=}5{+}1$)} \\
\cmidrule(lr){2-6} \cmidrule(lr){7-11}
& PSNR$\uparrow$ 
& RMSE$\downarrow$ 
& MAE$\downarrow$ 
& SSIM$\uparrow$
& LPIPS$\downarrow$
& PSNR$\uparrow$ 
& RMSE$\downarrow$ 
& MAE$\downarrow$ 
& SSIM$\uparrow$ 
& SAM$\downarrow$ \\
\midrule
w/o Resampling  
& 20.900 & 0.103 & 0.076 & 0.727 & 0.229
& 29.914 & 0.033 & 0.028 & 0.929 & 5.973 \\

w/o Correction  
& 20.899 & 0.103 & 0.077 & 0.726 & 0.228
& 29.891 & 0.035 & 0.031 & 0.918 & 5.991 \\

Mean-guided     
& 20.918 & 0.103 & 0.076 & 0.727 & 0.228
& 30.344 & 0.029 & 0.023 & 0.934 & 5.932 \\

Conv-guided     
& 20.931 & 0.103 & 0.076 & 0.729 & 0.228
& 31.135 & 0.028 & 0.024 & 0.937 & 5.891 \\

\textit{MAE}-guided
& \textbf{20.941} & \textbf{0.102} & \textbf{0.076} & \textbf{0.729} & \textbf{0.228}
& \textbf{31.234} & \textbf{0.028} & \textbf{0.023} & \textbf{0.937} & \textbf{5.885} \\
\bottomrule
\end{tabular}
\caption{Ablation study of different deterministic corrective resampling strategies.
Results are reported under fixed sampling configurations ($N{+}N^r{=}4{+}1$ for Sen2\_MTC\_New and $5{+}1$ for SEN12MS-CR-TS(EA)).}
\label{tab:resampling_strategy_ablation}
\end{table*}


\subsection{Computation Efficiency}
\label{sec:computation_efficiency}

\begin{table*}[t]
\centering
\setlength{\tabcolsep}{4pt}
\renewcommand{\arraystretch}{1.2}
\resizebox{\linewidth}{!}{
\begin{tabular}{lccccccccccccc}
\toprule
\textbf{Metric} 
& McGAN & Pix2Pix & AE & CycleGAN & STGAN & CTGAN & PMAA 
& CR-TS Net& UnCRtainTS & DDPM-CR & DiffCR & EMRDM & \M \\
\midrule
Params (M) 
& 54.42 & 11.45 & 6.53 & 11.44 & 231.93 & 642.92 & 3.44 
& 38.47 & 0.57 & 445.44 & 22.19 & 148.88 & 27.10 \\

MACs (G) 
& 35.88 & 29.47 & 17.78 & 56.80 & 1094.94 & 632.05 & 92.35 
& 7602.97 & 37.16 & 852.37 & 45.86 & 74.39 & 346.56 \\
Performance 
& 17.448 & 16.985 & 15.100 & 17.678 & 18.152 & 18.308 & 18.009 
& 15.347 & 18.770 & 18.742 & 19.150 & 20.249 & 20.941 \\
\bottomrule
\end{tabular}
}
\caption{Comparison of model complexity in terms of Parameters and MACs on the Sen2\_MTC\_New dataset.}
\label{tab:sen12ms_complexity}
\vspace{-2ex}
\end{table*}

\begin{figure}[htp]
    \centering
    \includegraphics[width=0.85\linewidth]{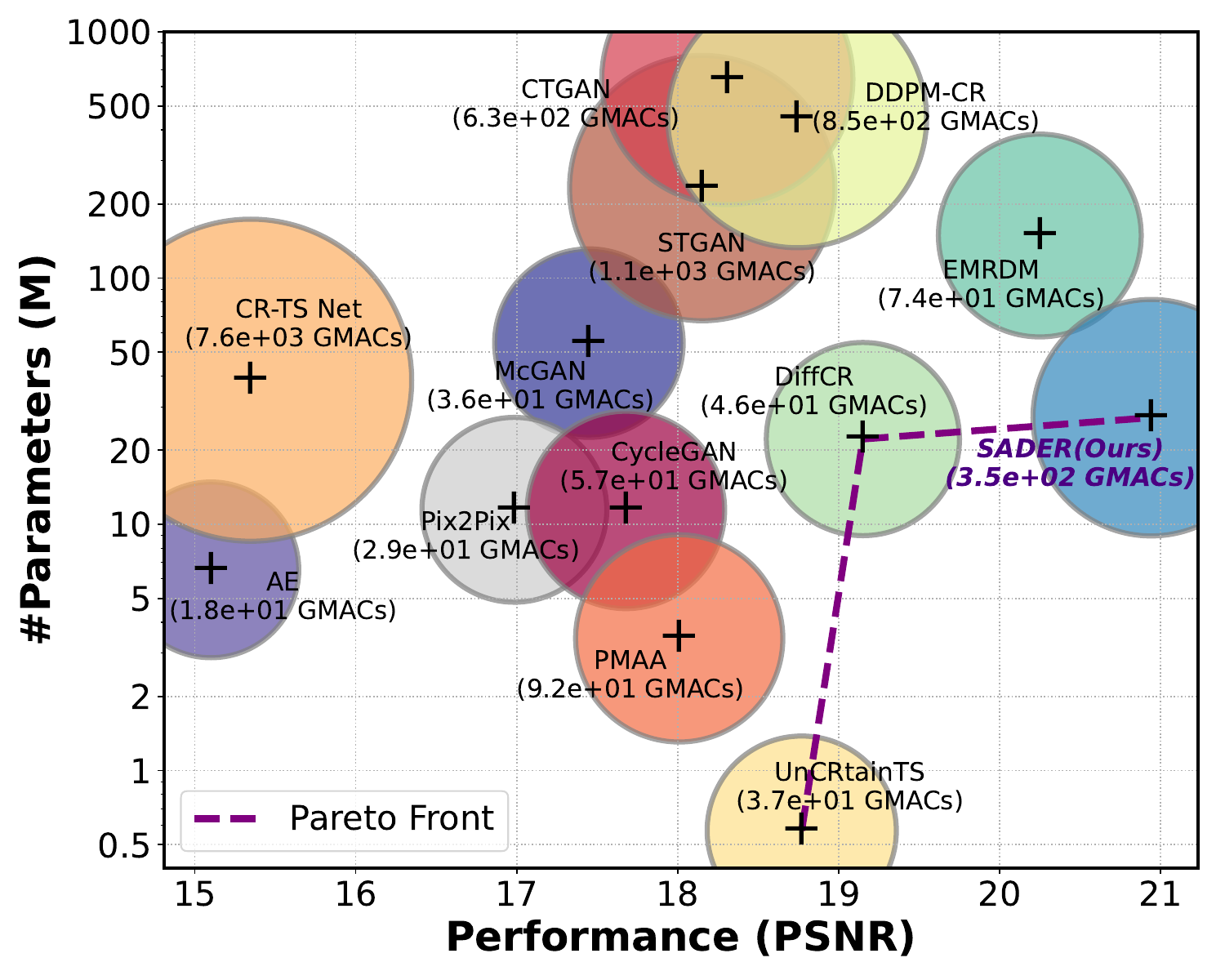}
    \caption{Trade-off between model performance (PSNR), parameter count, and inference cost (GMACs) on the Sen2\_MTC\_New dataset. The x-axis denotes PSNR (higher is better), the y-axis shows the number of parameters per source (in millions), and the marker size indicates GMACs. Purple dashed line represents the Pareto front. 
    }
    \vspace{-2ex}
    \label{fig:efficiency}
\end{figure}

From the perspective of practical multi-temporal cloud removal, computational efficiency is a critical consideration, as diffusion-based models often incur high inference costs that hinder large-scale deployment. As illustrated in Fig.~\ref{fig:efficiency} and Table~\ref{tab:sen12ms_complexity}, we evaluate \M against representative cloud removal approaches in terms of reconstruction performance (PSNR), parameter scale, and inference complexity measured by GMACs.

Efficiency-oriented models with compact architectures, such as DiffCR or UnCRtainTS, achieve relatively low inference cost but remain limited in reconstruction quality, whereas higher-capacity diffusion or temporal models, exemplified by DDPM-CR or CR-TS Net, improve restoration performance at the expense of substantially increased parameters or computational overhead. As shown in Fig.~\ref{fig:efficiency}, \M lies on the Pareto-optimal frontier, achieving the highest PSNR among all compared methods while maintaining moderate model size and inference cost. This balanced position between lightweight yet underperforming models and high-cost yet less deployable alternatives highlights \M as a more practical solution for large-scale multi-temporal remote sensing cloud removal.

\subsection{Effectiveness of Cloud-Aware Loss \& Deterministic Resampling}
\begin{figure}[t]
    \centering
    \includegraphics[width=\linewidth]{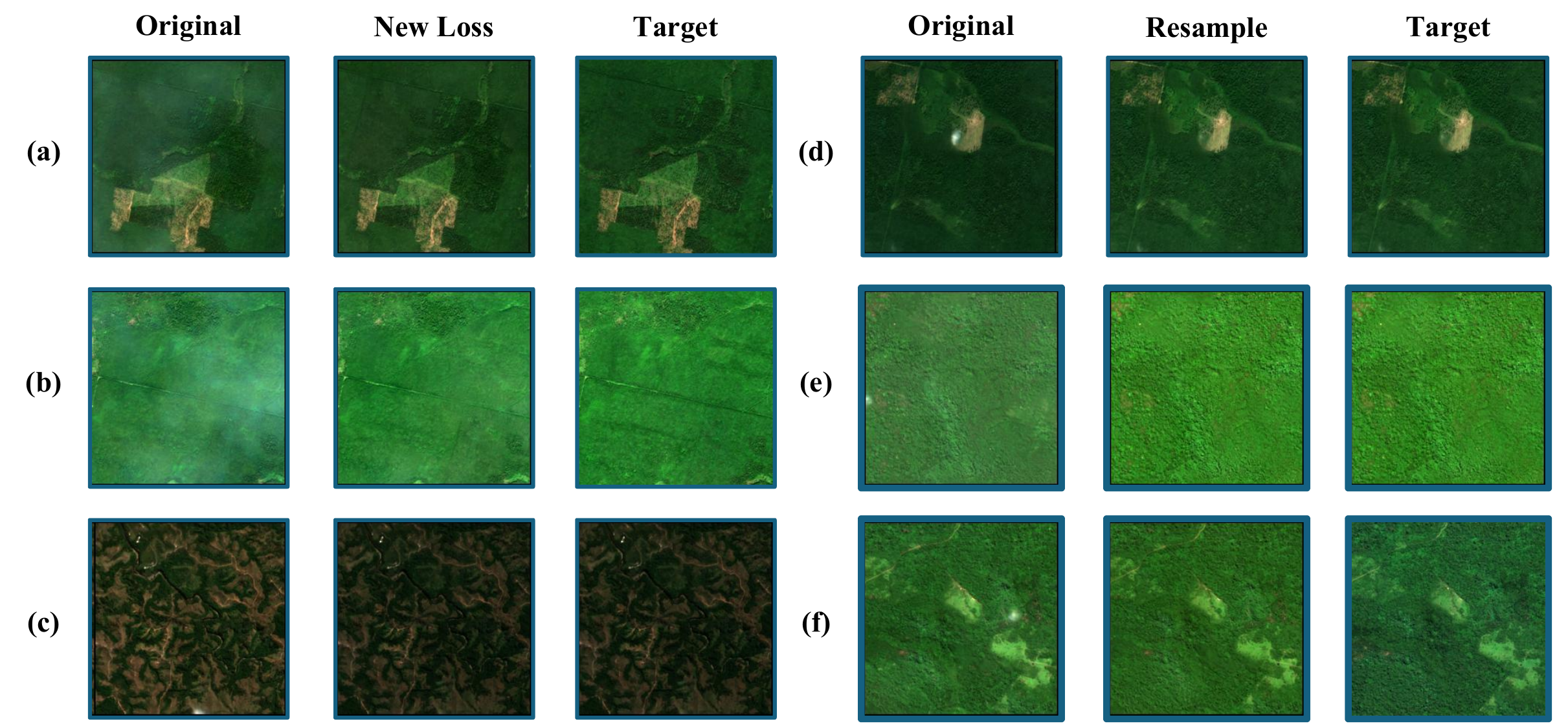}
    \caption{Qualitative analysis of the effect of the proposed cloud-aware loss and deterministic resampling strategy. 
    Columns (a)--(c) compare results produced by the original loss, the proposed cloud-aware loss, and the cloud-free target, respectively. Columns (d)--(f) show results before resampling, after guided resampling, and the target.
    }
    \label{fig:loss_resample}
    \vspace{-2ex}
\end{figure}

The effectiveness of the proposed cloud-aware loss is qualitatively illustrated in Fig.~\ref{fig:loss_resample}(a)--(c).
Compared with the original loss, the proposed formulation produces more faithful reconstructions under complex cloud-contaminated conditions. The proposed loss effectively suppresses residual cloud-related artifacts that remain in the results obtained using the original loss, leading to clearer surface structures and reduced cloud-like patterns.
In addition, the incorporation of the brightness consistency constraint alleviates intensity inconsistencies across the reconstructed regions, producing results that are more visually consistent with the cloud-free targets.
These observations indicate that jointly modeling cloud-dominated regions while enforcing brightness consistency improves both reconstruction accuracy and overall appearance consistency.

The effectiveness of the proposed deterministic resampling strategy is further illustrated in Fig.~\ref{fig:loss_resample}(d)--(f).
Compared with deterministic sampling without resampling, the proposed strategy effectively suppresses spurious high-intensity artifacts and small cloud-contaminated patterns inherited from the original cloudy observations. Deterministic resampling prevents unrealistic bright artifacts from being propagated through the sampling process, resulting in reconstructions that are more structurally consistent with the cloud-free targets.
These observations indicate that deterministic resampling functions as an effective refinement mechanism during inference, complementing the diffusion sampling process by selectively correcting unstable predictions.

\section{Conclusion}
In this work, we propose \M, a multi-temporal remote sensing cloud removal framework based on a mean-reverting diffusion model. 
By designing a lightweight network MTCDN, \M effectively models temporal dependencies across multi-temporal observations and leverages conditional information for guided reconstruction. 
Specifically, the proposed temporal fusion modules enable stable aggregation of cross-temporal information, while the hybrid attention mechanism enhances region-adaptive feature modeling, leading to more accurate and structurally consistent reconstruction results.
At the diffusion modeling level, \M introduces a cloud-aware training loss to guide region-focused learning under spatially varying uncertainty. 
Within this loss formulation, a brightness consistency constraint is incorporated to further stabilize the training process under multi-temporal conditions and to promote consistency in the reconstructed results. 
In addition, a deterministic resampling strategy is proposed to refine unreliable predictions under a fixed sampling budget. 
These designs jointly improve reconstruction accuracy and result stability in multi-temporal cloud removal tasks.
Extensive experiments on multiple multi-temporal datasets demonstrate that \M consistently outperforms existing SOTA methods across diverse scenarios, achieving notable improvements in spatial reconstruction fidelity and spectral consistency.

\bibliographystyle{IEEEtran}
\bibliography{ref_library}

\vspace{-5ex}

\begin{IEEEbiography}
{Yifan Zhang}
received the B.Eng. degree in Data Science and Big Data Technology at Tongji University in 2025. He is currently pursuing the Master of Data Science at the University of Michigan, Ann Arbor, with an expected graduation date of 2027. His research interests include computer vision, particularly diffusion models, and multimodal large language models.
\end{IEEEbiography}

\vspace{-5ex}

\begin{IEEEbiography}
{Qian Chen}
is an undergraduate student majoring in Data Science at the School of Computer Science and Technology, Tongji University, Shanghai, China, and is expected to graduate in 2026. His research interests include spatial-temporal data analysis, AI for Science applications, and interpretable deep learning.
\end{IEEEbiography}

\vspace{-5ex}

\begin{IEEEbiography}
{Yi Liu}is a second-year M.S. student in Computer Science and Technology at Tongji University’s ADMIS Lab, advised by Assoc. Prof. Wengen Li, expected to graduate in 2027. He received his B.E. in Computer Science and Technology from the College of Electronic and Information Engineering, Tongji University, in 2024. His research focuses on generative models, particularly diffusion models.
 
\end{IEEEbiography}

\vspace{-5ex}

\begin{IEEEbiography}
{Wengen Li}
received the B.Eng. degree and Ph.D. degree in Computer Science from Tongji University, Shanghai, China, in 2011 and 2017, respectively. In addition, he received a dual Ph.D. degree in Computer Science from the Hong Kong Polytechnic University in 2018. He is currently an Associate Professor of the School of Computer Science and Technology at Tongji University. His research interests include multi-modal artificial intelligence, and spatio-temporal intelligence for urban computing and ocean science. He is a member of China Computer Federation (CCF), a member of IEEE, and a member of ACM.
\end{IEEEbiography}

\vspace{-5ex}

\begin{IEEEbiography}
{Jihong Guan}
received the bachelor’s degree from Huazhong Normal University in 1991, the master’s degree from Wuhan Technical University of Surveying and Mapping (merged into Wuhan University since 2000), Wuhan, China, in 1998, and the PhD degree from Wuhan University, Wuhan, China, in 2002.
She is currently a professor of the Schoold of Computer Science and Technology, Tongji University, Shanghai, China. Before joining Tongji University, she served in the Department of Computer, Wuhan Technical University of Surveying  and Mapping from 1991 to 1997, as an assistant professor and an associate professor (since August 2000), respectively. She was an associate professor (2000-2003) and a professor (Since 2003) in the School of Computer, Wuhan University. Her research interests include databases, data mining, distributed computing, bioinformatics, and geographic information systems (GIS).
\end{IEEEbiography}

\vfill\textbf{}

\end{document}